\documentclass[11pt]{article}

\usepackage[final]{acl}
\usepackage{times}
\usepackage{latexsym}

\usepackage[T1]{fontenc}
\usepackage[utf8]{inputenc}

\usepackage{microtype}

\usepackage{inconsolata}

\usepackage{graphicx}

\usepackage{algorithm}
\usepackage{algorithmic}
\usepackage{enumitem}
\usepackage{amsmath}
\usepackage{titleref}
\usepackage{multirow}
\usepackage[most]{tcolorbox}
\usepackage{booktabs}
\usepackage{multirow}
\usepackage{amsmath,amssymb}
\usepackage{makecell}
\usepackage{enumitem}

\title{Worlds Within Words: Translating Culture in Ancient Chinese Texts with Multi-Agent Coordination}

\author{
 \textbf{Xiaoqi He\textsuperscript{1}},~~
 \textbf{Kaixin Lan\textsuperscript{1}},~~
 \textbf{Mu You\textsuperscript{2}},~~
 \textbf{Tao Fang\textsuperscript{2}},~~
 \textbf{Lidia S. Chao\textsuperscript{1}},~~
 \textbf{Derek F. Wong \textsuperscript{1}}\thanks{~Corresponding Author}
\\
\\
 \textsuperscript{1}NLP$^2$CT Lab, Department of Computer and Information Science, University of Macau
 \\
 \texttt{\{mc45283,lidiasc,derekfw\}@um.edu.mo, nlp2ct.kaixin@gmail.com}
 \\
 \textsuperscript{2}Institute of International Language Services Studies, Macau Millennium College
 \\
\texttt{youmuafonso@gmail.com, taofang@mmc.edu.mo}
}

\begin{document}
\maketitle
\begin{abstract}
Large language model (LLM)-based machine translation has advanced cross-cultural communication, yet it still struggles with culture-loaded words (CLWs) in ancient Chinese texts. The challenge extends beyond lexical alignment to deciding when and how culture-dependent knowledge should be explicated for readers lacking relevant background. Literal translation often preserves surface forms while missing underlying concepts, whereas over-explicitation harms conciseness and readability.
To address this problem, we formulate CLW translation as a selective explicitation task and propose \textbf{MACAT}, a \textbf{M}ulti-\textbf{A}gent \textbf{C}ulture-\textbf{A}ware \textbf{T}ranslation framework that dynamically identifies culturally salient phrases and injects concise explanatory knowledge when necessary. MACAT further incorporates a quality-aware reranking module for candidate selection and a multi-round evaluation agent that assesses translations across terminological precision, readability, fidelity, cultural preservation, and cultural explicitation.
Experiments on traditional Chinese medicine (TCM) classics and the \textit{Analects} show that, under a unified GPT-5.4 evaluation setting, MACAT consistently outperforms both the backbone model and general-purpose MT baselines on 100 TCM documents and a 20-chapter subset of the \textit{Analects}.
% Large language model (LLM)-based machine translation has significantly advanced cross-cultural communication, yet still struggles to recover culturally implicit relations in Classical Chinese culture-loaded words (CLWs). A key challenge lies not only in lexical alignment, but in determining when and how to explicitly surface culturally implicit knowledge for readers lacking sufficient cultural background. Literal translation often preserves surface form while missing underlying conceptual structures, whereas excessive explicitation harms conciseness and readability.
% To address this, we formulate CLW translation as a selective explicitation task and propose the MACAT framework, a \textbf{M}ulti-\textbf{A}gent \textbf{C}ulture-\textbf{A}ware \textbf{T}ranslation framework that dynamically identifies culturally triggered segments and injects concise explanatory knowledge accordingly. A quality-aware reranking module is further applied to select optimal candidates, and a multi-round evaluation agent is designed to assess translation quality across multiple dimensions, including terminological precision, readability, fidelity, cultural preservation, and cultural explicitation.
% % semantic fidelity, terminology accuracy, and implicit knowledge completion.
% Experiments on traditional Chinese medicine (TCM) classics and the \textit{Analects} show that, under a unified GPT-5.4 evaluation setting, the MACAT framework consistently outperforms both the backbone model and general-purpose MT baselines on 100 TCM documents and a 20-chapter subset of the \textit{Analects}.

\end{abstract}

% \noindent\textbf{Keywords:} Chinese culture-loaded words (CLWs); modular explicitation in translation; cross-cultural machine translation
% \vspace{0.8em}

\section{Introduction}
\begin{figure}[t]
  \centering
  \includegraphics[width=\columnwidth]{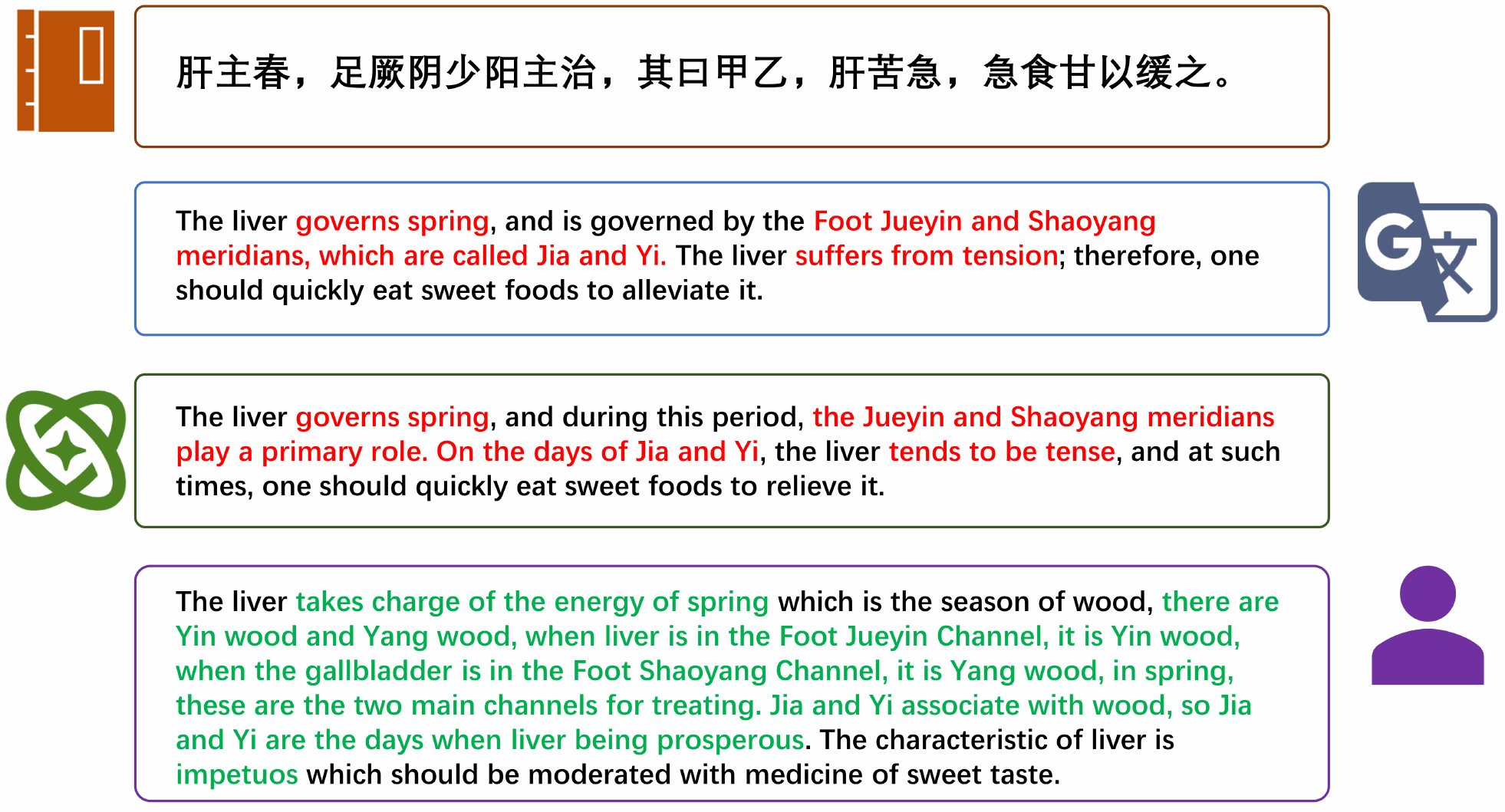}
  \caption{Translations for Chinese culture-loaded words from Google, LLM and human. Literal translations often preserve surface forms while failing to recover implicit cultural relations.}
  \label{fig:intro}
\end{figure}

One major obstacle in translating classical Chinese texts lies in culture-loaded words (CLWs). Unlike ordinary lexical items, many CLWs encode culturally grounded concepts through implicit epistemological structures rather than explicit surface meanings. Concepts such as ``yinyang'', bianzheng lunzhi'', and ``zangfu'' rely heavily on culturally shared background knowledge. 
As shown in Figure~\ref{fig:intro}, literal translations often fail to convey culturally embedded knowledge underlying CLWs to readers unfamiliar with Chinese culture \cite{yao-etal-2024-benchmarking,villacueva2025cammtbenchmarkingculturallyaware}.
% As illustrated in Figure~\ref{fig:intro}, literal and grammatically fluent translations often fail to recover these implicit conceptual relations for target readers who lack sufficient Chinese cultural background knowledge \cite{yao-etal-2024-benchmarking}. Consequently, preserving semantic adequacy in CLW translation requires more than surface-level lexical correspondence; it necessitates the selective reconstruction of culturally implicit meanings that are essential for comprehension.
% To address this issue, we formulate the problem as explicitation-oriented CLW translation, where only the minimal background knowledge required for understanding is selectively introduced into the translation process.
Existing automated translation methods provide limited support for this challenge. Document-level and LLM-based approaches mainly optimize semantic consistency and fluency \cite{wang-etal-2023-document-level,cui-etal-2024-efficiently,post2024escapingsentencelevelparadigmmachine}, while retrieval-augmented methods focus on terminology coverage and retrieval quality \cite{meng-etal-2022-fast,wang-etal-2022-efficient,conia-etal-2024-towards}. However, these systems rarely address two key questions in CLW translation: when cultural explicitation is necessary and how much explanatory knowledge should be injected. As a result, existing systems tend to either under-explain or over-elaborate \cite{freitag-etal-2022-results,zheng2023judging}.
To address these challenges, we propose \textbf{MACAT}, a fully inference-time Multi-Agent Culture-Aware Translation framework for explicitation-oriented translation. MACAT coordinates multiple agents to identify culturally sensitive phrases, construct compact knowledge cards, generate translation candidates under different knowledge-conditioning settings, and perform quality-aware reranking using reference-free quality estimation. Knowledge injection is applied only to culturally triggered segments, reducing unnecessary interference with translation. We further introduce a multidimensional evaluation agent that assesses translations across terminological precision, readability, fidelity, cultural preservation, and cultural explicitation.

\medskip
Our contributions are as follows:
\begin{itemize}%[nosep,leftmargin=*]
\item We formulate CLW translation as an explicitation-oriented translation task centered on selectively reconstructing culturally implicit knowledge.
\item We propose MACAT, a fully inference-time multi-agent framework that performs CLW extraction, knowledge injection, candidate generation, reranking, and consistency refinement without parameter fine-tuning.
\item We introduce a multidimensional evaluation agent tailored to explicitation-oriented CLW translation.
\item Experiments on TCM texts and the \textit{Analects} show that MACAT consistently outperforms strong LLM and general-purpose MT baselines.
\end{itemize}

\begin{figure*}[t]
\centering
\includegraphics[width=0.99\textwidth]{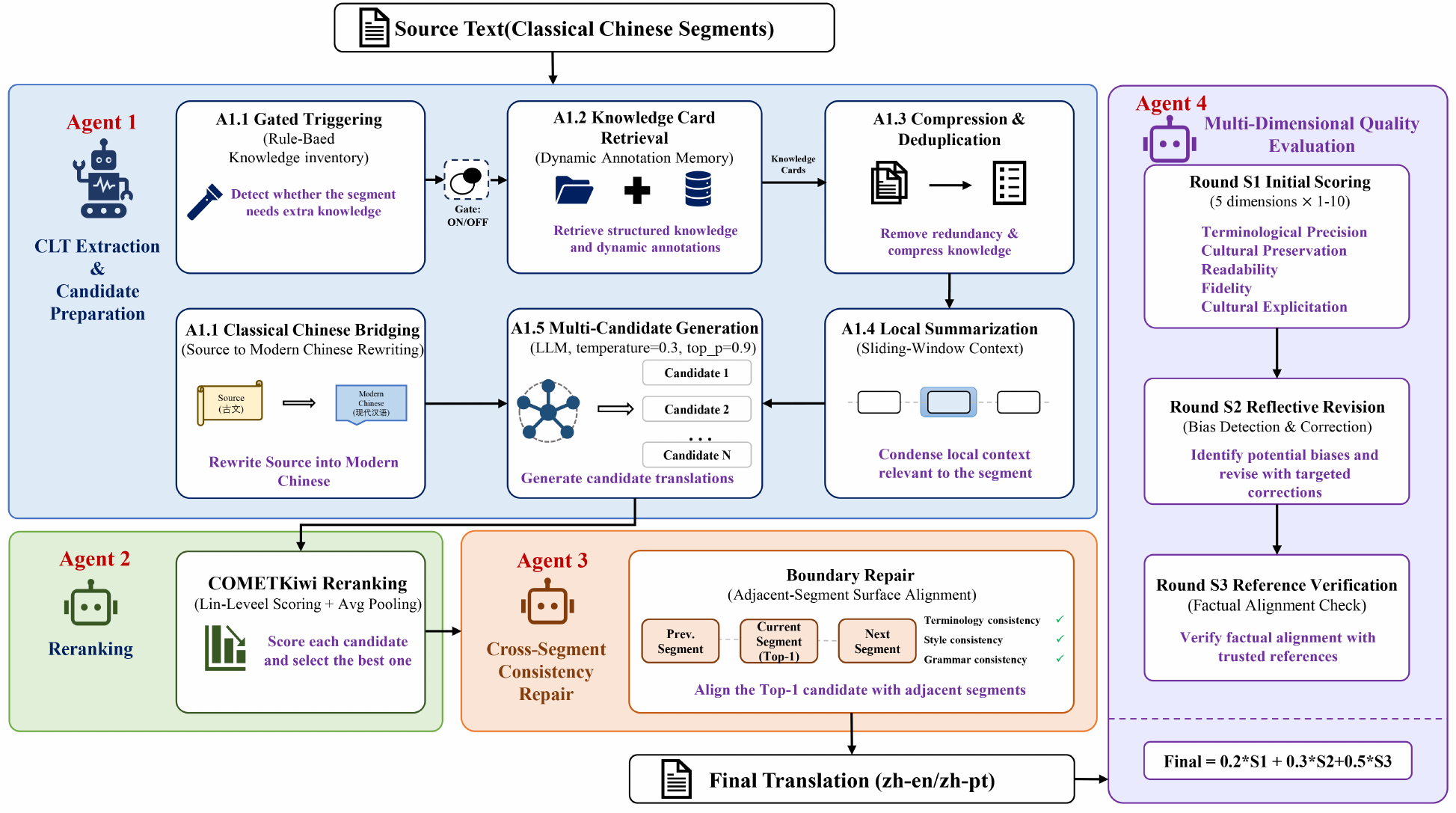}
\caption{Overview of MACAT. The framework consists of three main agents and an evaluation agent: (1) source preprocessing with classical Chinese bridging and local context summarization; (2) gated knowledge injection combining rule-based inventory lookup and dynamic annotation memory; (3) multi-candidate generation followed by COMETKiwi-based quality estimation reranking and boundary-aware local repair, (4) multi-dimensional translation quality evaluation.}
\label{fig:overview}
\end{figure*}

\section{Related Work}

\subsection{Constraint and Domain-aware Translation}

Constraint-based and domain-aware translation methods aim to improve terminology consistency and professional reliability in specialized domains. Representative approaches include terminology-constrained decoding \cite{ailem-etal-2021-encouraging}, translation-specific in-context learning \cite{li-etal-2023-mt2}, biomedical translation methods \cite{neves-etal-2023-findings, neves-etal-2024-findings}, and document-level LLM translation frameworks leveraging contextual prompting and long-context modeling \cite{wang-etal-2023-document-level, cui-etal-2024-efficiently, hu-etal-2024-large-language}. Retrieval-enhanced MT methods further incorporate external memory or nearest-neighbor retrieval to improve domain adequacy and contextual consistency \cite{khandelwal2021nearestneighbormachinetranslation, meng-etal-2022-fast, wang-etal-2022-efficient}. Recent work has begun to explore the dynamic nature of knowledge injection at inference time \cite{li-etal-2025-leveraging}, finding that retrieving domain-relevant exemplars (few-shot) is more effective than generating terminological definitions. However, this work primarily focuses on the source of knowledge rather than the timing and necessity of explicitation. 
Conversely, we treat Chinese culture-loaded word translation as a problem of selective explicitation, where cultural knowledge is activated only when additional interpretation becomes necessary.
%In contrast, our work formulates Chinese culture-loaded word (CLW) translation as a selective explicitation problem, dynamically activating culturally grounded knowledge only when additional interpretation is necessary.

\subsection{Multi-Agent Frameworks for Translation}

LLM-based translation research increasingly adopts modular and agent-style inference frameworks that decompose translation into retrieval, planning, refinement, or verification stages. Representative studies include Chain-of-Dictionary Prompting (CoD) \cite{lu-etal-2024-chain}, contextual refinement and LLM post-editing frameworks \cite{koneru-etal-2024-contextual}, retrieval-augmented translation methods based on multilingual knowledge graphs \cite{conia-etal-2024-towards}, and document-level modular prompting methods for discourse-aware translation \cite{wang-etal-2023-document-level, cui-etal-2024-efficiently}. Recent reflective refinement frameworks further explore iterative self-correction for translation improvement. TEaR \cite{feng-etal-2025-tear} introduces a Translate--Estimate--Refine paradigm using MQM-style feedback for iterative refinement. While such frameworks improve translation quality through multi-stage inference, they mainly optimize fluency and coherence, with limited modeling of culture-aware explicitation. In contrast, MACAT explicitly centers the framework around CLW-aware explicitation control and knowledge-guided translation refinement.

\subsection{Evaluation and LLM-as-a-Judge}

Recent research has shown that lexical-overlap metrics such as BLEU are insufficient for evaluating semantically rich and culturally grounded translation phenomena \cite{freitag-etal-2022-results}. Neural evaluation frameworks such as COMET and COMETKiwi provide stronger semantic adequacy estimation \cite{rei-etal-2022-cometkiwi}, while LLM-as-a-Judge methods further enable rubric-based and reflective assessment using large language models \cite{liu-etal-2023-g, zheng2023judging}. Recent studies also reveal important limitations of fine-tuned judge models, including weak generalizability, evaluator bias, and poor aspect-specific calibration across evaluation settings \cite{huang-etal-2025-empirical}. These limitations become particularly critical for explicitation-oriented CLW translation, where terminology precision, cultural fidelity, implicit knowledge completion, and over-explicitation correspond to distinct evaluation dimensions. However, translation quality hinges on meeting specific specifications \cite{feng-etal-2025-tear}. Therefore, we adopt a multi-round multidimensional evaluation Agent with cross-judge validation and reflective reassessment to improve robustness and evaluation reliability.

\section{Method}

% \noindent We propose a fully inference-time framework for explicitation-oriented translation of Chinese culture-loaded words (CLWs). The architecture is organized as a multi-agent framework centered on CLW extraction and knowledge-informed candidate preparation. Quality-aware reranking, cross-segment consistency repair, and multi-dimensional evaluation are incorporated as complementary stages to improve translation reliability and coherence. The framework adopts a profile-based configuration in which each profile controls the activation and capacity of auxiliary components, enabling deployment from lightweight direct translation to a full framework with bridging and boundary repair. The current implementation supports zh$\rightarrow$en and zh$\rightarrow$pt translation directions. A structural overview is presented in Figure~\ref{fig:overview}.
\noindent As illustrated in Figure~\ref{fig:overview}, we propose a fully inference-time framework for explicitation-oriented translation of Chinese culture-loaded words (CLWs). The framework is formulated as a multi-agent architecture centered on CLW extraction and knowledge-guided candidate generation. To enhance translation fidelity and discourse-level coherence, the framework further incorporates quality-aware reranking, cross-segment consistency refinement, and multi-dimensional evaluation as auxiliary optimization agents. To support flexible deployment under different computational and performance requirements, the framework adopts a profile-driven configuration mechanism, where each profile specifies the activation and capacity of auxiliary modules, ranging from lightweight direct translation to the complete framework with semantic bridging and boundary-repair components. 
% The current implementation supports zh$\rightarrow$en and zh$\rightarrow$pt translation directions. 
% An overview of the overall architecture is illustrated in Figure~\ref{fig:overview}.

\subsection{CTW Extraction \& Knowledge-Informed Candidate Preparation}

The key challenge in CLW translation is not knowledge retrieval itself, but determining when explicitation is necessary and how much cultural information should be injected. We address this problem through a gated knowledge construction.

\paragraph{CLW-triggered Gating.} \textbf{Agent 1} first applies a lightweight detection gate to identify segments potentially requiring cultural explicitation. Detection is performed using rule patterns covering major CLW categories, including yinyang theory, five elements, zang-fu systems, meridians, heavenly stems, six climatic pathogens, and Confucian terminology.
If no trigger is detected, the segment bypasses all knowledge-related modules and is translated directly by the backbone LLM. 
% This design reduces unnecessary processing and suppresses over-explicitation.

\paragraph{Structured Knowledge Construction.} For triggered segments, the MACAT framework activates a structured knowledge inventory represented as JSON-based rules(see Appendix~\ref{sec:prompt-knowledge} for the knowledge extraction prompt template). Each rule contains four components: trigger words, conceptual relations, application constraints, and a generation template. For example, the heavenly stem Jia activates associations with the Wood element, Liver system, yang polarity, and corresponding meridian pathways. Activated rules are instantiated into compact cultural knowledge cards used during translation prompting. 
To avoid excessive injection, broad conceptual families are filtered using keyword-overlap scoring, and only the highest-relevance entries are retained.

\paragraph{Dynamic CLW Annotation.}  
Beyond predefined glossaries, we additionally employ LLM-based CLW annotation (see Appendix~\ref{sec:prompt-annotation}). Using a dedicated prompting strategy, the backbone model automatically identifies culturally implicit concepts and produces concise explanatory notes for downstream translation.
Annotation is repeated three times with high-temperature sampling. Only words consistently detected across runs are retained through majority voting. This mechanism improves recall while reducing unstable annotations.

\paragraph{Knowledge Budgeting \& Context Conditioning.}
Rule-based knowledge cards and LLM-generated annotations are first merged via semantic deduplication and then filtered under a profile-dependent knowledge budget. 
% When the total injected content exceeds the budget, lower-confidence or weakly supported explanations are removed.
For triggered segments, the agent optionally introduces two auxiliary contextual signals:
(1) \textbf{Local summarization}, which compresses surrounding segments into a brief context summary to preserve topical continuity and prior explicitation (see Appendix~\ref{sec:prompt-context});
(2) \textbf{Classical Chinese bridging}, which rewrites Classical Chinese into modern Chinese while preserving original CLWs (see Appendix~\ref{sec:prompt-context}).
Both components serve to enhance contextual disambiguation, without introducing additional cultural knowledge.

\paragraph{Multi-Path Candidate Generation.} 
% When the detection gate is disabled, the framework operates as a single-pass translation pipeline. When the gate is enabled, the framework generates multiple candidates by varying two factors: the source form and the injected knowledge context. Source forms include the original classical Chinese segment and the modern Chinese bridge output, while knowledge conditions include no auxiliary knowledge, knowledge cards only, and knowledge cards with contextual summaries. When both bridging and summarization are enabled, the system produces up to six candidates per segment.

% All candidates are generated using the same decoding parameters. The prompt structure is fixed across all candidates(see Appendix~\ref{sec:prompt-translation}, Tables~\ref{tab:prompt-trans-en}). It begins with the task description, followed by auxiliary blocks in a predefined order, and ends with the source segment. This design ensures that differences among candidates arise from input variation rather than prompt formatting.
% When the detection gate is disabled, the agent reduces to a standard single-pass translation without auxiliary cultural processing. When enabled, it performs multi-path candidate generation by varying both source representation and injected cultural context. 
The source representation is either the original Classical Chinese segment or its modern Chinese bridging paraphrase, while the knowledge configuration includes three settings: no auxiliary knowledge, knowledge cards only, and knowledge cards with contextual summaries. Under the full configuration, the \textbf{Agent 1} generates up to six candidates per segment.
To ensure controlled comparison, all candidates are produced with identical decoding settings and a unified prompt template (see Appendix~\ref{sec:prompt-translation}). The prompt consists of a fixed instruction followed by optional auxiliary blocks in a predefined order, ending with the source segment. This design ensures that differences among outputs stem from input variations rather than prompt formulation.

\begin{table*}[ht]
\centering
\resizebox{.99\textwidth}{!}{
    \begin{tabular}{lccc}
    \toprule
    Corpus & Documents & Segments & Description \\
    \midrule
    PES & 100 & 1{,}005 &
    Five TCM subdomains with reference translations \\
    SES & 50 & 333 &
    Reference-free ranking for generalization analysis \\
    \textit{Analects} subset & 20 & 477 &
    Supplementary examples of Confucian culture-loaded words \\
    \bottomrule
    \end{tabular}
}
    \caption{Experimental corpora. PES is Primary Evaluation Set, SES is Supplementary Evaluation Set.}
    \label{tab:data}
\end{table*}

\subsection{Quality-Driven Reranking}

%\textbf{Agent 1} focuses on preparing knowledge-informed inputs for CLW-bearing segments. Translation generation and candidate selection are handled by a separate system that produces multiple candidates under different input conditions. 

\paragraph{COMETKiwi-Based Reranking.} Then \textbf{Agent 2} ranks them using a reference-free quality estimator. For a segment $s_i$ with candidate set $\{c_i^{(1)}, \dots, c_i^{(M)}\}$, each candidate translation is evaluated line by line using the wmt22-cometkiwi-da checkpoint of COMETKiwi~\cite{rei-etal-2022-cometkiwi}. The evaluation produces per-line quality scores $q_{i,j}^{(m)}$, which are aggregated into segment-level scores through average pooling after alignment with the source segment:

\begin{equation}
s(c_i^{(m)})=\frac{1}{|L_i|}\sum_{j=1}^{|L_i|} q_{i,j}^{(m)}.
\label{eq:candidate_score}
\end{equation}
The objective of reranking is not to identify a universally “best” translation, but to balance two competing risks: filtering out candidates with insufficient knowledge injection while suppressing outputs whose explanations are excessively verbose and therefore detrimental to translation conciseness. Local summarization and classical Chinese bridging do not inherently guarantee improvements, but they expand the candidate space available for this trade-off.

\subsection{Cross-Segment Consistency Repair}

Since translation is performed independently at the segment level, inconsistencies may arise across segment boundaries, including terminology drift, stylistic shifts, and redundant explicitation.
To address this, \textbf{Agent 3} optionally applies a boundary-aware repair. For each adjacent segment pair, a local context window is constructed, and only boundary-adjacent text is rewritten while interior content remains unchanged (see Appendix~\ref{sec:prompt-context}).
This agent introduces no additional knowledge and does not alter prior detection results, serving solely to improve local coherence and consistency.

\subsection{Multi-Dimensional Translation Quality Evaluation}

% We evaluate explicitation-oriented CLW translation using a three-round multi-dimensional protocol designed to measure both translation quality and explicitation behavior. System outputs are evaluated across five dimensions: Terminological Precision, Syntactic Fluency
% , Cultural Fidelity, Literal Information Completeness, Implicit Knowledge Explicitation. Details of the evaluation Criteria are provided in Appendix~\ref{app:evaluation}

% \paragraph{Three-Round Evaluation:}
% Round 1 performs independent scoring without reference translations; Round 2 applies reflective reassessment to reduce halo bias, scale inconsistency, and length bias; Round 3 conducts reference-assisted factual verification when expert translations are available.

% Final scores are computed through weighted aggregation:
% \begin{equation}
% \text{Final} = 0.2S^{(1)} + 0.3S^{(2)} + 0.5S^{(3)},
% \label{eq:final_score}
% \end{equation}
% This pipeline enables finer-grained distinction between terminology errors, missing cultural knowledge, and over-explicitation than single-score evaluation schemes.
We evaluate explicitation-oriented CLW translation using a three-round multidimensional agent covering five dimensions: Terminological Precision, Readability, Fidelity, Cultural Preservation, and Cultural Explicitation (see Appendix~\ref{app:evaluation}).
% We evaluate explicitation-oriented CLW translation using a three-round multidimensional agent covering five dimensions: Terminological Precision, Syntactic Fluency, Cultural Fidelity, Literal Information Completeness, and Implicit Knowledge Explicitation (see Appendix~\ref{app:evaluation}).

\paragraph{Three-Round Evaluation.}
Round 1 performs reference-free scoring; Round 2 applies reflective reassessment to mitigate halo bias, scale inconsistency, and length bias; Round 3 conducts reference-assisted factual verification when expert translations are available.
Final scores are computed as:
\begin{equation}
\text{Final} = 0.2S^{(1)} + 0.3S^{(2)} + 0.5S^{(3)}
\label{eq:final_score}
\end{equation}
The protocol enables fine-grained analysis of terminology errors, missing cultural knowledge, and over-explicitation beyond single-score evaluation.

\section{Experiment}

\subsection{Benchmark}

% We evaluate the proposed framework on TCM and \textit{Analects} Chinese culture-loaded translation settings. The primary benchmark consists of a frozen snapshot of parallel Corpus, covering five TCM subdomains: acupuncture, background, disease, herb, and treatment. The dataset contains 100 documents and 1,005 aligned segments with expert reference translations. Primary Evaluation Set is used for main system comparison, ablation analysis, and reference-assisted third-round evaluation.

% To evaluate generalization under reference-free conditions, we additionally construct a supplementary dataset containing 50 documents and 333 segments drawn from the same five TCM subdomains. Supplementary Evaluation Set does not provide expert references and is evaluated only using the first two rounds of the proposed evaluation pipeline.

% To test transferability beyond medical text, we further evaluate the framework on a 20-chapter subset of the \textit{Analects}, containing 477 segments with Confucian culture-loaded terminology. The subset is used in the zh$\rightarrow$pt experiments\cite{hu-etal-2024-large-language}.

% Dataset statistics are summarized in Table~\ref{tab:data}.
We evaluate the proposed framework on TCM and \textit{Analects} CLW translation tasks. The primary benchmark consists of a frozen parallel corpus covering five TCM subdomains: acupuncture, background, disease, herb, and treatment. It contains 100 documents and 1,005 aligned segments with expert reference translations, and is used for main experiments, ablation studies, and third-round reference-assisted evaluation.
To assess generalization under reference-free conditions, we additionally construct a supplementary TCM dataset containing 50 documents and 333 segments from the same subdomains. Evaluation on this set uses only the first two rounds of the proposed protocol.
To evaluate cross-domain transferability, we further test the framework on a 20-chapter subset of the \textit{Analects}, comprising 477 segments with Confucian culture-loaded terminology for zh$\rightarrow$pt translation~\cite{hu-etal-2024-large-language}.
Dataset statistics are summarized in Table~\ref{tab:data}.

%%%  main reslut %%%%%%%%%%%%%%%%%%%%%%%%%%%%%%%%%%%%%%%%%%%%%%%%%%%%%
%  Terminological Precision, Readability, Fidelity, Cultural Preservation, and Cultural Explicitation
% Terminological Precision, Syntactic Fluency, Cultural Fidelity, Literal Information Completeness, and Implicit Knowledge Explicitation
\begin{table*}[htp]
\centering
%\small
\footnotesize
\resizebox{\textwidth}{!}{
\begin{tabular}{lccccccccc} 
\toprule
Method & $S^{(1)}$ & $S^{(2)}$ & $S^{(3)}$ & Final & \begin{tabular}[c]{@{}c@{}}Termino-\\logy\end{tabular} & \begin{tabular}[c]{@{}c@{}}Reada-\\bility\end{tabular} & Fidelity & \begin{tabular}[c]{@{}c@{}}Cultural\\Preservation\end{tabular} & \begin{tabular}[c]{@{}c@{}}Cultural\\Explicitation\end{tabular} \\ 
% Method & $S^{(1)}$ & $S^{(2)}$ & $S^{(3)}$ & Final & \begin{tabular}[c]{@{}c@{}}Termino-\\logy\end{tabular} & Fluency & \begin{tabular}[c]{@{}c@{}}Cultural\\imagery\end{tabular} & \begin{tabular}[c]{@{}c@{}}Literal\\compl.\end{tabular} & \begin{tabular}[c]{@{}c@{}}Implicit\\knowl.\end{tabular} \\ 
\midrule
Google & 5.136 & 4.738 & 6.567 & 5.732 & 5.218 & 7.220 & 5.063 & 6.077 & 3.790 \\ 
CoD & 6.081 & 5.972 & 7.415 & 6.715 & 6.359 & 8.220 & 5.853 & 6.797 & 4.384 \\
Tower-plus & 5.814 & 5.237 & 5.960 & 5.714 & 5.675 & 6.172 & 5.569 & 6.293 & 4.642 \\ 
\midrule
LLM & \multicolumn{1}{l}{} & \multicolumn{1}{l}{} & \multicolumn{1}{l}{} & \multicolumn{1}{l}{} & \multicolumn{1}{l}{} & \multicolumn{1}{l}{} & \multicolumn{1}{l}{} & \multicolumn{1}{l}{} & \multicolumn{1}{l}{} \\
%-Qwen3-4B-Instruct & 5.460 & 4.592 & 6.812 & 5.876 & - & - & - & - & - \\
-Qwen3-8B & 5.450 & 5.021 & 6.535 & 5.864 & 5.540 & 6.516 & 5.425 & 7.292 & 3.569 \\
-Qwen3.5-plus & 6.650 & 5.887 & 7.795 & 6.993 & 6.405 & 7.409 & 6.223 & 7.648 & 4.535 \\
-Deepseek-V3 & 6.764 & 6.481 & 7.511 & 7.053 & 7.089 & 7.978 & 6.620 & 7.945 & 4.960 \\ 
\midrule
MACAT &  &  &  &  &  &  &  &  &  \\
%-Qwen3-4B-Instruct & 5.892 & 5.176 & 6.952 & 6.207 & - & - & - & - & - \\
-Qwen3-8B & 5.580 & 5.105 & 7.581 & 6.438 & 5.959 & 7.284 & 5.619 & 6.885 & 4.692 \\
-Qwen3.5-plus & \textbf{7.613} & \textbf{7.164} & 8.022 & 7.683 & 7.299 & 7.691 & \textbf{7.253} & \textbf{8.469} & \textbf{7.286} \\
-Deepseek-V3 & 7.483 & 7.094 & \textbf{8.254} & \textbf{7.752} & \textbf{7.443} & \textbf{8.437} & 7.132 & 8.276 & 6.730 \\
\bottomrule
\end{tabular}
}
\caption{Main results under GPT-5.4-based evaluation. $S^{(1)}$, $S^{(2)}$, and $S^{(3)}$ are  dimension scores per round; five dimension columns report the average across all three evaluation rounds. MACAT, Deepseek-V3, and Google scores are computed on the full 120-document evaluation set. \textbf{Bold} indicates the highest Final score in each column.}
\label{tab:main-results}
\end{table*}
%%%%%%%%%%%%%%%%%%%%%%%%%%%%%%%%%%%%%%%%%%%%%%%%%%%%%%%%

\subsection{Experimental Setup}
\paragraph{Translation Backbones.}
% The framework operates entirely at inference time and does not fine-tune any model parameters. Four LLM backbones are used for translation generation: Qwen3-4B-Instruct, Qwen3-8B, Qwen3.5-Plus, DeepSeek-V3. 
% The framework operates entirely at inference time without any parameter fine-tuning. 
We experimented with four large language models as candidate backbone models for translation generation, including 
%Qwen3-4B-Instruct,
Qwen3-8B, Qwen3.5-Plus \footnote{\url{https://github.com/QwenLM}}, and DeepSeek-V3 \cite{deepseekai2025deepseekv3technicalreport}. Based on overall translation quality, DeepSeek-V3 was selected as the backbone model for the rest experiments.
The direct backbone output without the proposed framework is reported as the LLM baseline.
% For each backbone, the proposed framework generates translation candidates under different knowledge-conditioning settings and reranks them using quality estimation. The direct backbone output without the proposed pipeline is reported as the LLM baseline.

\paragraph{Baselines.} We compare MACAT against four external methods:

\begin{itemize}[nosep,leftmargin=*]
\item Google: commercial machine translation;
\item CoD: prompt-optimized translation baseline\cite{lu-etal-2024-chain};
\item Tower-plus: translation-specialized multilingual LLM baseline\cite{rei2025towerbridginggeneralitytranslation};
\item LLM: direct backbone translation without the Multi-Agent;
\end{itemize}

\paragraph{Quality Estimation \& Reranking.}Candidate reranking uses the wmt22-cometkiwi-da checkpoint of COMETKiwi in a reference-free setting. The reranker assigns line-level quality scores, which are aggregated into segment-level scores for final candidate selection.

% \subsection{Evaluation methods}

% \paragraph{Zh$\rightarrow$Pt Extension}
% To examine cross-lingual transferability, we further construct a zh$\rightarrow$pt translation setting on a 20-chapter subset of the \textit{Analects}. This setting follows the same pipeline used in the zh$\rightarrow$en experiments, with only the target-language instructions, Portuguese stylistic constraints, and Chinese--Portuguese terminology resources replaced.

% \begin{table}[t]
% \centering
% \begin{tabular}{p{0.23\columnwidth}p{0.67\columnwidth}}
% \toprule
% Item & Configuration \\
% \midrule
% GPU & NVIDIA RTX 4090D $\times 1$ (24GB) \\
% CPU & 15 vCPU Intel(R) Xeon(R) Platinum 8474C \\
% CUDA & $\leq$ 12.4 \\
% Python & 3.8.10 \\
% % Core software & LangChain/OpenAI-compatible API, \newline COMETKiwi, pandas, openpyxl, python-docx \\
% Inference setting & API-based generation with local COMETKiwi reranking \\
% \bottomrule
% \end{tabular}
% \caption{Experimental environment.}
% \label{tab:env}
% \end{table}

\subsection{Main Results}

Table~\ref{tab:main-results} reports the main results under GPT-5.4 evaluation across commercial MT systems, prompt-based baselines, direct backbone translation, and MACAT instantiated with different LLM backbones. Overall, MACAT consistently outperforms all baselines across nearly all evaluation dimensions.
Among all variants, MACAT with DeepSeek-V3 achieves the highest overall Final score (7.752), significantly surpassing direct DeepSeek-V3 (7.053), CoD (6.715), Google Translate (5.732), and Tower-plus (5.714). The improvements remain stable across all three evaluation rounds, indicating consistent gains rather than metric-specific overfitting. We attribute this to structured cultural reasoning and controlled explicitation.
We also observe a backbone-dependent pattern: MACAT-Qwen3.5-plus achieves the best scores on several individual dimensions, including Cultural Fidelity (7.253), Cultural Preservation (8.469), and Cultural Explicitation (7.286), suggesting that explicitation capability is not model-specific. However, DeepSeek-V3-based MACAT achieves the best overall performance and the most stable cross-round results, indicating a better global balance and robustness. These findings suggest complementarity between backbone capacity and the proposed cultural control mechanism.
Across dimensions, the largest gains consistently appear in culturally sensitive aspects. Compared to direct DeepSeek-V3, MACAT improves Cultural Explicitation (4.960$\rightarrow$6.730) and Cultural Fidelity (6.620$\rightarrow$7.132), addressing the core challenge of CLW translation. In contrast, Readability improves marginally (7.978$\rightarrow$8.437), reflecting that modern LLMs already exhibit strong surface-level fluency.
Finally, MACAT improves completeness without sacrificing readability. Cultural Preservation increases from 7.945 to 8.276 while maintaining competitive Readability. This is consistent with its design, where knowledge injection is selectively applied to CLW spans and over-explicit candidates are filtered during reranking, preserving naturalness while enhancing cultural grounding.

\section{Analysis}
\subsection{Human Evaluation}
To validate the reliability of the proposed automatic evaluation pipeline, we additionally conduct human evaluation under the same criteria. The evaluation set consists of three zh$\rightarrow$en TCM translations (avg. 1,300 words) and two zh$\rightarrow$pt \textit{Analects} translations (avg. 900 words), annotated by three domain researchers for each translation direction.
As shown in Table~\ref{tab:human-evaluation}, MACAT achieves the best overall performance across all CLW-related dimensions, supporting the reliability of the automatic evaluation protocol. MACAT obtains slightly lower readability and fidelity scores than direct DeepSeek-V3 translation, which we attribute to the inherent trade-off introduced by explicit cultural clarification.
% , where selective explicitation of culture-loaded concepts may moderately reduce stylistic conciseness while substantially improving cultural comprehensibility.

\begin{table*}[!ht]
\centering
% \resizebox{.75\textwidth}{!}{
\begin{tabular}{lccccc} 
\toprule
Model & Terminology & Readability & Fidelity & \begin{tabular}[c]{@{}c@{}}Cultural\\Preservation\end{tabular} & \begin{tabular}[c]{@{}c@{}}Cultural\\Explicitation\end{tabular} \\ 
\midrule
Google & 3.800 & 5.200 & 3.800 & 3.933 & 2.733 \\
Deepseek-V3 & 7.067 & \textbf{7.867} & \textbf{7.533} & 7.600 & 6.533 \\
MACAT & \textbf{7.867} & 7.800 & 7.467 & \textbf{8.467} & \textbf{8.000} \\
\bottomrule
\end{tabular}
% }
\caption{Human evalution on MACAT, Deepseek-V3 and Google. MACAT improves scores of Terminology, Cultural Preservation and Cultural Explicitation, with only minor reductions in Readability and fidelity. \textbf{Bold} indicates the highest score in each dimension.}
\label{tab:human-evaluation}
\end{table*}

\subsection{Case Study}
Through multi-agent coordination, MACAT generates translations that are more conceptually informative and interpretively accessible. Its main advantage lies in recovering implicit knowledge embedded in Classical Chinese texts, which are often difficult even for contemporary Chinese readers to fully interpret.
% 肝苦急，急食甘以缓之
A representative example is the phrase ``gan ku ji, ji shi gan yi huan zhi'' from \textit{Huangdi Neijing}. We compare six translations produced under different ablation settings (see Appendix \ref{Appendix:Case Study}). While the baseline systems mainly rely on literal translation, MACAT reconstructs the implicit therapeutic reasoning encoded in the source text. This difference is particularly evident in the translation of ``gan'', which baseline systems render as ``sweet foods'' or ``sweet flavors'', reducing a technical TCM category to ordinary dietary language. In contrast, MACAT contextualizes sweetness within TCM flavor theory, where flavors function as therapeutic categories associated with specific physiological effects.
The original sentence compresses pathology, diagnosis, and treatment rationale into a highly elliptical structure. Although the baseline translations preserve lexical brevity, they also obscure the underlying therapeutic logic. MACAT selectively expands the sentence to make this logic explicit: pathological liver excess or constraint is moderated through sweet-flavored substances. This explicitation improves readability and knowledge transfer for readers unfamiliar with TCM theory.

% \subsection{Cross-Judge Robustness Analysis}

% To examine the robustness of the evaluation results, we further evaluate the same outputs using three different LLM judges: GPT-5.4, Claude, and Gemini. Table~\ref{tab:different-judge-results} reports the Final scores using DeepSeek-V3 as the translation backbone.
% Across all judges, MACAT consistently remains the top performer. More importantly, the relative performance gap between MACAT and the direct backbone stays stable despite noticeable calibration differences among evaluators. Under GPT-5.4, MACAT improves over DeepSeek-V3 by +0.699, while the margins under Claude and Gemini are +0.395 and +0.377, respectively. Although the absolute gains vary, all three judges produce the same ordering: MACAT $>$ DeepSeek-V3 $>$ Google.

% Overall, the results indicate that the improvements introduced by MACAT are robust across evaluators rather than artifacts of a specific LLM judge.

% %%%%%%%%%%%%%%%%%%%%%%%%%%%%%%%%%%%%%%%
% \begin{table}[!h]
%     \centering
%     \resizebox{\linewidth}{!}{
%     \begin{tabular}{lccc}
%         \toprule
%         Judge & Deepseek-V3 & Google & MACAT \\
%         \midrule
%         GPT-5.4 & 7.053 & 5.732 & 7.752 \\
%         Claude & 7.127 & 6.014 & 7.522 \\
%         Gemini & 7.281 & 5.386 & 7.658 \\
%         \bottomrule
%     \end{tabular}
%     }
%     \caption{Final scores under different judge models with DeepSeek-V3 as the translation backbone.}
%     \label{tab:different-judge-results}
% \end{table}
%%%%%%%%%%%%%%%%%%%%%%%%%%%%%%%%%%%%

\subsection{Transferability Beyond TCM}
To evaluate generalization beyond TCM discourse, experiments are conducted on a 20-chapter subset of the \textit{Analects} containing Confucian culture-loaded terminology. The zh$\rightarrow$pt results further demonstrate cross-lingual transferability. As shown in Table~\ref{tab:zhpt-lunyu}, MACAT reaches a Final score of 7.317, outperforming DeepSeek-V3 (6.955) and Google (3.941). The strongest gains appear in CLW-related dimensions: CLW accuracy improves by +0.230 relative to DeepSeek-V3, while Explicitation-related performance increases by +1.023. Fluency differences remain comparatively small because both systems already generate relatively natural Portuguese outputs.
These findings suggest that the framework captures general explicitation principles rather than memorizing domain-specific terminology patterns.

\begin{table*}[!ht]
\centering
%\small
% \resizebox{.8\textwidth}{!}{
\begin{tabular}{lcccccc}
\toprule
Model & Final & \makecell{Terminology} & \makecell{Readability} & Fidelity & \makecell{Cultural\\Preservation} & \makecell{Cultural\\Explicitation} \\ 
\midrule
Google & 3.941 & 2.986 & 5.515 & 2.959 & 3.242 & 2.493 \\
Deepseek-V3 & 6.955 & 6.762 & 8.312 & 6.466 & 7.901 & 4.489 \\
MACAT & \textbf{7.317} & \textbf{6.992} & \textbf{8.465} & \textbf{6.790} & \textbf{7.948} & \textbf{5.512} \\
\bottomrule
\end{tabular}
% }
\caption{Zh$\rightarrow$Pt results on the \textit{Analects} subset (lunyu:c1-c20). MACAT also achieves the best performance on the zh$\rightarrow$pt translation task. \textbf{Bold} indicates the highest score in each dimension.}
\label{tab:zhpt-lunyu}
\end{table*}

\subsection{Ablation Study}
To analyze component contributions, we remove key modules from MACAT and report averaged results in Table~\ref{tab:ablation-results}.
Knowledge-related components contribute most to performance. Removing RAG/Notes causes the largest drop in the final score (7.752$\rightarrow$7.089, -0.663), with major degradation in Cultural Preservation and Cultural Explicitation (-0.658 and -0.690), indicating that external knowledge primarily improves recovery of culturally implicit information. A similar trend appears in the w/o Replace Law setting, where domain-aligned CLW knowledge is replaced with unrelated legal-domain knowledge. Although performance remains above direct translation, the final score decreases to 7.216, with the largest loss in Cultural Explicitation (-0.847), while Readability remains stable (8.100).
By contrast, stabilization modules produce smaller but consistent gains. Removing summary compression reduces the Final score by 0.470, mainly affecting Fidelity and Cultural Preservation, while removing MC Bridge causes a smaller decline (-0.358) concentrated in Cultural Explicitation and Fidelity. These results suggest that bridge and summary modules mainly improve contextual coherence rather than serving as primary knowledge sources.
Across all variants, Readability remains relatively stable (7.883$\rightarrow$8.437), whereas Cultural Explicitation and Fidelity show the largest fluctuations. Overall, the ablation results reveal a clear functional decomposition: knowledge injection drives most improvements, while bridge and summary modules mainly stabilize contextual consistency and coherence.
\begin{table*}[!ht]
\centering
% \resizebox{.5\textwidth}{!}{
\begin{tabular}{lcccccc}
\toprule
Variant & Final & \makecell{Terminology} & \makecell{Readability} & Fidelity & \makecell{Cultural\\Preservation} & \makecell{Cultural\\Explicitation} \\ 
\midrule
MACAT & \textbf{7.752} & \textbf{7.443} & \textbf{8.437} & \textbf{7.132} & \textbf{8.276} & \textbf{6.730} \\
w/o MC Bridge & 7.394 & 7.157 & 8.250 & 6.750 & 8.119 & 6.168 \\
w/o Summary & 7.282 & 6.935 & 8.196 & 6.630 & 7.934 & 6.153 \\
w/o RAG/Notes & 7.089 & 6.884 & 7.883 & 6.616 & 7.618 & 6.040  \\
w/o Replace Law & 7.216 & 7.085 & 8.100 & 6.743 & 7.786 & 5.883 \\
\bottomrule
\end{tabular}
% }
\caption{Ablation results. When key modules are removed, scores of every dimensions decrease. Each domain is computed on the c1--c10 subset translated with Deepseek-V3. \textbf{Bold} indicates the highest score in each dimension.}
\label{tab:ablation-results}
\end{table*}
\subsection{Reference-Free Translation}
\begin{table*}[!ht]
\centering
% \resizebox{.9\textwidth}{!}{
\begin{tabular}{lccccccc}
\toprule
System & $S^{(1)}$ & $S^{(2)}$ & \makecell{Terminology} & \makecell{Readability} & Fidelity & \makecell{Cultural\\Preservation} & \makecell{Cultural\\Explicitation} \\
\midrule
MACAT & \textbf{7.474} & \textbf{7.123} & \textbf{7.113} & \textbf{8.103} & \textbf{6.718} & \textbf{8.132} & \textbf{6.499} \\
Deepseek-V3 & 6.722 & 6.436 & 6.665 & 7.882 & 6.088 & 7.808 & 4.447 \\
Google & 5.874 & 5.391 & 5.328 & 7.424 & 5.144 & 6.576 & 3.692 \\
\bottomrule
\end{tabular}
% }
\caption{Results on RandomCorpus without using reference text. This supplementary set only uses the first two rounds, so we report $S^{(1)}$, $S^{(2)}$, and dimension means instead of Final. \textbf{Bold} indicates the highest score in each column.}
\label{tab:random-results}
\end{table*}

To further evaluate robustness beyond the main benchmark, we conduct experiments on RandomCorpus, a reference-free dataset constructed from randomly sampled passages in classical Chinese texts, including \textit{Shanghan Lun}, \textit{Bencao Gangmu}, and excerpts from \textit{Journey to the West}. Evaluation is performed using only the first two assessment rounds, without reference-assisted verification.
As shown in Table~\ref{tab:random-results}, MACAT consistently outperforms both the backbone model and Google across all dimensions. The largest gain is observed in Cultural Explicitation (+2.052, 4.447$\rightarrow$6.499), followed by improvements in Terminological Precision (+0.448) and Cultural Preservation (+0.324).
These improvements remain stable without reference-based evaluation, indicating that gains stem from improved recovery of implicit cultural knowledge during generation rather than reference alignment.Fluency shows relatively smaller improvements, suggesting that the framework primarily enhances culturally grounded explicitation rather than surface-level naturalness.

\section{Conclusion}
In this work, we formulate Chinese culture-loaded word (CLW) translation as an explicitation-oriented translation task that selectively reconstructs culturally implicit knowledge for target-side comprehension. To address this challenge, we propose MACAT, a fully inference-time Multi-Agent Culture-Aware Translation Framework that integrates CLW detection, knowledge injection, candidate generation, reranking, and consistency refinement through coordinated agent-based control without parameter fine-tuning.
We further introduce a multidimensional evaluation benchmark and criteria tailored to explicitation-oriented CLW translation, enabling fine-grained evaluation of Terminological Precision, Readability, Fidelity, Cultural Preservation, and Cultural Explicitation.
Experiments on both TCM corpora and the \textit{Analects} demonstrate that MACAT consistently outperforms strong LLM baselines and general-purpose machine translation systems. 
The results further validate the effectiveness, robustness and generalizability of inference-time multi-agent coordination for culturally aware translation. 
% The results of human evaluation further validate the effectiveness, robustness and generalizability of inference-time multi-agent coordination for culturally aware translation. 
Additionally, the automatic evaluation results are largely consistent with human evaluation, further supporting the effectiveness of the proposed evaluation agent.
We hope this work provides a useful foundation for future research on culturally informed machine translation and the international dissemination of classical Chinese knowledge systems.

% This paper formulates CLW translation as controlled knowledge injection at inference time. The system injects short explanations only for triggered segments, while bridge, summary, reranking, and local repair remain auxiliary mechanisms.

% On the main TCM set, the shared 50-document subset, and the \textit{Analects}, the full system stays above direct model translation and generic MT baselines. The clearest improvements appear in terminology accuracy, literal completeness, and implicit knowledge completion. The zh$\rightarrow$pt results on the \textit{Analects} are consistent with these differences.

%\clearpage
\section{Limitation \& Future Work}
The current framework has four main limitations. First, the knowledge interface is still dominated by hand-written rules and templates. This design gives precise control, but rule coverage and maintenance cost limit direct scaling to broader genres and larger corpora. Second, the main results rely heavily on LLM-as-a-Judge evaluation. Multi-round scoring and cross-judge checks reduce variance, but alignment with human evaluation still requires more validation. Third, the cross-lingual evidence is incomplete, especially for zh$\rightarrow$pt, where the current results demonstrate transferability more clearly than comprehensive comparative coverage. Fourth, multi-path generation and COMETKiwi reranking increase inference cost, which remains a practical constraint for large-scale or latency-sensitive deployment.

Future work should focus on coverage, validation, and efficiency. On the modeling side, the knowledge interface can be extended with automatically induced rules, lightweight retrieval, or knowledge-graph-backed mappings to reduce manual engineering. On the evaluation side, human assessment, significance testing, and finer-grained error attribution are needed to calibrate automatic judges on controlled explicitation. On the transfer side, the framework should be tested on more language pairs and more text types, including poetry, military writings, late imperial fiction, and historical prose, to measure robustness beyond the current TCM and \textit{Analects} settings \cite{conia-etal-2024-towards}.

\section*{Acknowledgements}
This work was supported in part by the Science and Technology Development Fund of Macau SAR (Grant Nos. FDCT/0007/2024/AKP, EF2024-00185-FST), the UM and UMDF (Grant Nos. MYRG-GRG2024-00165-FST-UMDF, MYRG-GRG2025-00236-FST), the Tencent AI Lab Rhino-Bird Research Program (Grant No. EF2023-00151-FST), the Dr. Stanley Ho Medical Development Foundation (Grant No. SHMDF-AI/2026/001), and the National Natural Science Foundation of China (Grant No. 62266013). This work was performed in part at SICC which is supported by SKL-IOTSC, and HPCC supported by ICTO of the University of Macau.

% \clearpage

\bibliography{custom.bib}

%\clearpage
\appendix
\section*{Appendix}
\section{Additional Analysis}
\label{Appendix:Additional Results}

\subsection{Case Study}
\label{Appendix:Case Study}

To further understand the contribution of individual components in MACAT, Table~\ref{tab:tcm_case} presents a qualitative case study under different ablation settings. The example contains several Traditional Chinese Medicine (TCM) concepts, including \textit{Liver}, \textit{Qi}, and the therapeutic role of \textit{sweet flavor}, which require cultural and domain-specific interpretation beyond literal translation.

The table compares the full MACAT system with several controlled variants obtained by removing individual modules. By examining the resulting translations, we can directly observe how different knowledge sources and contextual mechanisms influence the explicitation of culturally grounded meanings.

The full MACAT system generates the most informative interpretation, explicitly rendering the pathological state of the Liver as energy becoming ``impetuous or constrained'' and explaining the therapeutic function of sweet-flavored foods within the framework of flavor theory. In contrast, removing knowledge-related components often leads to partial loss of cultural meaning. For example, variants without knowledge augmentation tend to simplify the underlying TCM concepts into generic descriptions such as ``tension,'' ``rigidity,'' or ``dislike urgency,'' thereby weakening the connection to the original medical theory.

This example illustrates that the improvements achieved by MACAT do not arise from surface-level paraphrasing alone. Instead, they stem from the integration of structured cultural knowledge, contextual reasoning, and translation-time explicitation, which together enable more faithful rendering of culturally embedded concepts in Classical Chinese medical texts.

\begin{table*}
\centering
\begin{tabular}{lp{12cm}} 
\toprule
\textbf{Version} & \textbf{Translation}    \\ 
\midrule
MACAT            & The Liver suffers when its energy becomes impetuous or constrained; therefore, one should quickly consume foods with a Sweet taste, which are therapeutically used to relax and tonify based on flavor theory to moderate this urgency.  \\
w/o RAG/Notes    & The Liver suffers from tension and rigidity; therefore, one should promptly consume sweet flavors to relax and harmonize it.                                                                                                             \\
w/o MC Bridge    & The liver dislikes tension, so one should quickly eat sweet foods to relieve it.                                                                                                                                                         \\
w/o Summary      & The liver tends to be tense, and at such times, one should quickly eat sweet foods to relieve it.                                                                                                                                        \\
w/o Replace Law  & The liver dislikes urgency, and one should eat sweet foods to alleviate it.                                                                                                                                                              \\
Deepseek-V3      & The liver dislikes urgency, so one should quickly eat sweet foods to soothe it.                                                                                                                                                          \\
\bottomrule
\end{tabular}
\caption{Translations of ``gan ku ji, ji shi gan yi huan zhi'' under different ablation settings. `` Deepseek-V3'' denotes translation generated without any auxiliary enhancement techniques.}
\label{tab:tcm_case}
\end{table*}

\subsection{Cross-Judge Robustness Analysis}

To examine the robustness of the evaluation results, we further evaluate the same outputs using three different LLM judges: GPT-5.4, Claude, and Gemini. Table~\ref{tab:different-judge-results} reports the Final scores using DeepSeek-V3 as the translation backbone.
Across all judges, MACAT consistently remains the top performer. More importantly, the relative performance gap between MACAT and the direct backbone stays stable despite noticeable calibration differences among evaluators. Under GPT-5.4, MACAT improves over DeepSeek-V3 by +0.699, while the margins under Claude and Gemini are +0.395 and +0.377, respectively. Although the absolute gains vary, all three judges produce the same ordering: MACAT $>$ DeepSeek-V3 $>$ Google.

Overall, the results indicate that the improvements introduced by MACAT are robust across evaluators rather than artifacts of a specific LLM judge.

%%%%%%%%%%%%%%%%%%%%%%%%%%%%%%%%%%%%%%%
\begin{table}[!ht]
    \centering
    \resizebox{\linewidth}{!}{
    \begin{tabular}{lccc}
        \toprule
        Judge & Deepseek-V3 & Google & MACAT \\
        \midrule
        GPT-5.4 & 7.053 & 5.732 & 7.752 \\
        Claude & 7.127 & 6.014 & 7.522 \\
        Gemini & 7.281 & 5.386 & 7.658 \\
        \bottomrule
    \end{tabular}
    }
    \caption{Final scores under different judge models with DeepSeek-V3 as the translation backbone.}
    \label{tab:different-judge-results}
\end{table}

% \begin{table*}[!ht]
% \centering
% % \resizebox{.8\textwidth}{!}{
% \begin{tabular}{lcccccc}
% \toprule
% Model & Final & \makecell{Terminology} & \makecell{Readability} & Fidelity & \makecell{Cultural\\Preservation} & \makecell{Cultural\\Explicitation} \\ 
% \midrule
% Google & 3.941 & 2.986 & 5.515 & 2.959 & 3.242 & 2.493 \\
% Deepseek-V3 & 6.955 & 6.762 & 8.312 & 6.466 & 7.901 & 4.489 \\
% MACAT & \textbf{7.317} & \textbf{6.992} & \textbf{8.465} & \textbf{6.790} & \textbf{7.948} & \textbf{5.512} \\
% \bottomrule
% \end{tabular}
% % }
% \caption{Zh$\rightarrow$Pt results on the \textit{Analects} subset (lunyu:c1-c20). \textbf{Bold} indicates the highest score in each dimension.}
% \label{tab:zhpt-lunyu}
% \end{table*}

% \begin{table*}[!ht]
% \centering
% % \resizebox{.75\textwidth}{!}{
% \begin{tabular}{lccccccc}
% \toprule
% System & $S^{(1)}$ & $S^{(2)}$ & \makecell{Terminology} & \makecell{Readability} & Fidelity & \makecell{Cultural\\Preservation} & \makecell{Cultural\\Explicitation} \\
% \midrule
% MACAT & \textbf{7.474} & \textbf{7.123} & \textbf{7.113} & \textbf{8.103} & \textbf{6.718} & \textbf{8.132} & \textbf{6.499} \\
% Deepseek-V3 & 6.722 & 6.436 & 6.665 & 7.882 & 6.088 & 7.808 & 4.447 \\
% Google & 5.874 & 5.391 & 5.328 & 7.424 & 5.144 & 6.576 & 3.692 \\
% \bottomrule
% \end{tabular}
% % }
% \caption{Results on RandomCorpus. This supplementary set only uses the first two rounds, so we report $S^{(1)}$, $S^{(2)}$, and dimension means instead of Final. \textbf{Bold} indicates the highest score in each column.}
% \label{tab:random-results}
% \end{table*}

\section{Detailed Evaluation Criteria}
\label{app:evaluation}
This study adopts a five-dimensional evaluation framework, in which each dimension is assigned an integer score ranging from 1 to 10, accompanied by a brief English justification. The average score across the five dimensions constitutes the overall score for Round 1, denoted as $S^{(1)}$. After reflective revision in Round 2, the updated score is denoted as $S^{(2)}$. Following factual verification against the reference translation in Round 3, the final revised score is denoted as $S^{(3)}$. The overall final score is computed as:
\[
\text{Final} = 0.2S^{(1)} + 0.3S^{(2)} + 0.5S^{(3)}.
\]
All evaluation dimensions are equally weighted, and each evaluator independently completes the full assessment across all dimensions.

\subsection{Terminological Precision}

\textbf{Evaluation Objective:} This dimension assesses whether culturally specific terminology is translated in a professional and standardized manner while accurately conveying its underlying cultural connotations. For Traditional Chinese Medicine (TCM) corpora, particular attention is paid to whether the translations preserve the logical relationships embedded in the yin--yang and five-elements framework (e.g., distinctions between yin/yang wood and yin/yang meridians). For non-medical corpora, the evaluation focuses on whether the translated terminology maintains consistency within the corresponding philosophical.

\textbf{Scoring Criteria:}
\begin{itemize}[nosep,leftmargin=*]
\item \textbf{Score 1:} Core terminology is entirely mistranslated or omitted. A typical error would be translating ``Zu Jueyin'' literally as foot negative, thereby completely losing the concept of the meridian system.
\item \textbf{Score 5:} Most key words are translated correctly, but certain expressions remain non-standardized, or the corresponding yin--yang / Heavenly Stems relationships are insufficiently specified. For example, the terminology may be translated correctly while failing to distinguish the associated five-elements attributes.
\item \textbf{Score 10:} All terminology adopts widely recognized English equivalents and accurately conveys the internal logical relationships embedded in the terms. For instance, distinctions such as Yin Wood versus Yang Wood are explicitly preserved together with their associated meridians, and lexical choices appropriately reflect the cultural connotations of the source terminology (e.g., translating ``shenxinjuzai'' as harmonize the body and spirit rather than relying on a literal rendering).
\end{itemize}

\subsection{Cultural Preservation}

\textbf{Evaluation Objective:} This dimension evaluates whether the translation preserves the cultural connotations and underlying cognitive framework of the source text. Particular attention is given to whether core Traditional Chinese Medicine (TCM) philosophical principles---such as ``correspondence between humans and nature'' (\textit{tian ren xiang ying}), ``following the rhythms of the four seasons'' (\textit{fa si shi}), and ``analogical reasoning through symbolic correspondence'' (\textit{qu xiang bi lei})---are retained, rather than replaced or filtered through modern cultural interpretations.

\textbf{Scoring Criteria:}
\begin{itemize}[nosep,leftmargin=*]
\item \textbf{Score 1:} The cultural connotations are entirely replaced or filtered through modern conceptual interpretations. A representative error would be translating ``fa sishi'' as seasonal treatment, thereby losing the philosophical meaning of ``fa'' as ``to follow'' or ``to model oneself after.''
\item \textbf{Score 5:} The basic meaning is conveyed, but the philosophical expressions are simplified or blurred. The underlying cultural logic is weakened, although not fundamentally distorted. For example, the lexical choices may fail to fully reflect the conventional semantic nuances of TCM discourse.
\item \textbf{Score 10:} The translation fully preserves and accurately conveys distinctive cultural reasoning frameworks such as ``correspondence between humans and nature'' and ``analogical classification through symbolic correspondence.'' Lexical choices (e.g., harmonize, relieve) facilitate readers’ understanding of this unique epistemological system while maintaining the original worldview centered on the integration of humanity and nature.
\end{itemize}

\subsection{Readability}

\textbf{Evaluation Objective:} This dimension evaluates whether the translation is natural and fluent, conforms to academic writing conventions, and is free from conspicuous translationese or fragmented grammatical structures.

\textbf{Scoring Criteria:}
\begin{itemize}[nosep,leftmargin=*]
\item \textbf{Score 1:} Frequent grammatical errors and severely fragmented sentence structures substantially hinder comprehension.
\item \textbf{Score 5:} The grammar is generally correct, but the translation exhibits noticeable translationese and lacks natural stylistic fluency.
\item \textbf{Score 10:} The writing is idiomatic and fluent, with accurate logical transitions and stylistic conventions consistent with academic discourse, showing no evident traces of literal translation.
\end{itemize}

\subsection{Fidelity}

\textbf{Evaluation Objective:} This dimension assesses whether the translation fully preserves the literal information contained in the source text, without omitting or improperly removing critical cultural facts, relationships, conditions, or modifying elements.

\textbf{Scoring Criteria:}
\begin{itemize}[nosep,leftmargin=*]
\item \textbf{Score 1:} Severe omissions are present, with more than half of the key information from the source text missing.
\item \textbf{Score 5:} Most explicit information is preserved, but certain cultural facts, conditional relationships, or modifying elements are omitted.
\item \textbf{Score 10:} All literal information is comprehensively preserved, including critical cultural facts, relationships, conditions, and restrictive modifiers.
\end{itemize}

\subsection{Cultural Explicitation}

\textbf{Evaluation Objective:} This dimension assesses whether the translation provides only the minimal implicit cultural background knowledge necessary for target-language readers to comprehend classical texts. Such supplementation should be accurate, concise, and conducive to understanding, while avoiding both insufficient clarification and excessive explanation.

\textbf{Scoring Criteria:}
\begin{itemize}[nosep,leftmargin=*]
\item \textbf{Score 1:} No attempt is made to address implicit knowledge that is essential for target readers’ comprehension, resulting in a translation that may appear fluent but fails to convey the underlying cultural logic meaningfully.
\item \textbf{Score 5:} Necessary clarification is provided only intermittently or inconsistently; alternatively, the supplementary explanations may be accurate yet insufficiently focused, leading to redundant elaboration.
\item \textbf{Score 10:} Implicit knowledge is supplemented only where necessary, and the additions are accurate, concise, and effective in facilitating reader comprehension. No unnecessary explanation is imposed on explicitly stated information.
\end{itemize}

\section{Summary of Evaluation Instructions Across Rounds}

\subsection{Round 1: Initial Multi-Dimensional Evaluation}

\textbf{Primary Objective:} Without access to any reference translation, evaluators assign independent scores (integer values from 1 to 10) for each of the five evaluation dimensions based solely on the source text and the translation. For each dimension, a justification of no more than 12 English words must be provided. The output must consist exclusively of a single JSON object.

\textbf{Procedure:}
\begin{enumerate}[nosep,leftmargin=*]
\item The system injects the source text (source) and translation (translation) into the evaluation prompt as inputs.
\item Evaluators act in the role of ``experts in the translation evaluation of classical Chinese cultural texts'' and assess the following five dimensions individually: Terminological Precision, Cultural Fidelity, Syntactic Fluency, Literal Information Completeness, and Implicit Knowledge Explicitation.
\item Each dimension is assigned an integer score on a six-level scale ranging from 1 to 10, accompanied by a concise English justification containing no more than 12 words.
\item The output format must strictly follow JSON syntax, with the following English field names: terminology\_accuracy, grammar\_readability, cultural\_imagery, literal\_information\_completeness, and implicit\_knowledge\_explicitation.
\end{enumerate}

\subsection{Round 2: Reflective Review and Score Revision}

\textbf{Primary Objective:} Based on the Round 1 evaluation results, reviewers conduct a reflective reassessment to identify potential issues such as the Halo Effect, Length Bias, Scale Mismatch, inconsistencies across evaluation dimensions, or deviations from academically accepted standards. Scores are then revised or confirmed accordingly.

\textbf{Procedure:}
\begin{enumerate}[nosep,leftmargin=*]
\item The system reinjects the complete JSON output from Round 1 (prev\_json), together with the source text and translation, into the evaluation prompt.
\item Evaluators carefully examine the following potential issues:
  \begin{itemize}[nosep,leftmargin=*]
  \item \textbf{Halo Effect:} Whether overall fluency has caused specific terminological errors to be overlooked;
  \item \textbf{Length Bias:} Whether longer translations are unfairly penalized or shorter translations unfairly rewarded;
  \item \textbf{Scale Mismatch:} Whether scoring criteria are applied consistently across all translations;
  \item \textbf{Cross-Dimensional Inconsistency:} Whether scores across different dimensions are logically inconsistent (e.g., low readability but exceptionally high cultural fidelity);
  \item \textbf{Academic Consistency:} Whether judgments regarding terminology and cultural interpretation align with established academic conventions.
  \end{itemize}
\item Evaluators output the revised scores together with revision rationales. If no revision is necessary, the original scores are confirmed and reissued unchanged.
\end{enumerate}

\textbf{Output Requirement:} The response must contain only a JSON object identical in structure to that of Round 1. Field names and formatting must remain unchanged, and no additional explanatory text is permitted.

\subsection{Round 3: Reference-Based Factual Verification}

\textbf{Primary Objective:} This round incorporates an expert reference translation for factual verification in order to determine whether additional explanatory content in the evaluated translation constitutes value-added elaboration or merely redundant paraphrasing, while avoiding the assumption that the reference translation represents the upper bound of acceptable expression.

\textbf{Procedure:}
\begin{enumerate}[nosep,leftmargin=*]
\item In addition to the source text and evaluated translation, the system injects an expert reference translation (reference), explicitly labeled as ``not a standard answer, but provided solely for factual verification.''
\item Evaluators use the reference translation exclusively for the following two forms of assessment:
  \begin{itemize}[nosep,leftmargin=*]
  \item \textbf{Semantic Alignment:} Whether the evaluated translation conveys the same core cultural facts as the reference translation;
  \item \textbf{Value-Added Analysis:} Whether explanatory additions beyond the reference translation constitute necessary and accurate value-enhancing clarification or merely redundant paraphrasing.
  \end{itemize}
\item Scoring follows the principles below:
  \begin{itemize}[nosep,leftmargin=*]
  \item If the evaluated translation is more explicit than the reference translation, no penalty should be assigned provided that the cultural facts remain accurate and the additions are necessary and restrained;
  \item If the evaluated translation is shorter and more fluent but makes comprehension more difficult for readers lacking cultural background knowledge, it should not receive a higher score;
  \item Evaluation should continue to prioritize the objectives of explicitation-oriented translation rather than similarity to the wording of the reference translation.
  \end{itemize}
\end{enumerate}

\textbf{Output Requirement:} The response must contain only a JSON object whose structure and field names are identical to those used in the previous two rounds. No additional explanatory text is permitted.

\section{Example of a Logical Reasoning Chain}

The following example presents the first valid rule in the knowledge base (See Table~\ref{tab:rule-stem-element}, Stem\_Element\_Organ\_Link, illustrating in full its triggering conditions, logical mapping relationships, application rules, and output template.

\begin{table*}[t]
\centering\small
\begin{tabular}{p{0.20\textwidth}p{0.70\textwidth}}
\toprule
Property & Description \\
\midrule
Rule Name & Stem\_Element\_Organ\_Link \\
Triggers & Jia, Yi \\
\midrule
\multicolumn{2}{c}{\textbf{Logic Chain}} \\
\midrule
Jia  & Yang Wood; associated five-elements attribute is Wood, corresponding to the Liver organ; linked to the Foot Shaoyang Gallbladder Channel. \\
Yi  & Yin Wood; associated five-elements attribute is Wood, corresponding to the Liver organ; linked to the Foot Jueyin Liver Channel. \\
\midrule
Application Rule & When Jia or Yi appears, explicate its association with the Wood element and the Liver organ, distinguishing the corresponding meridians according to yin--yang attributes. \\
\bottomrule
\end{tabular}
\caption{Specification of the Stem\_Element\_Organ\_Link rule.}
\label{tab:rule-stem-element}
\end{table*}

\subsection{Output Template}
% \begin{verbatim}
the days of {Stem} (which correspond to {element}
element and govern the {organ}; {Stem} is {nature}
associated with {property})
% \end{verbatim}

\textbf{Example Template Instantiation:} Taking ``Jia'' as an example, the actual generated output is:
The days of Jia (which correspond to the Wood element and govern the Liver; Jia is Yang Wood associated with the Wood element and the prosperity of the Liver).

\vspace{0.8em}
\section{Systems and Controlled Variants}
\vspace{-0.3em}

To facilitate reproducibility and clarify the role of each component in MACAT, Table~\ref{tab:systems} summarizes all systems and controlled variants used throughout the experiments. The table includes both external baselines and internal ablation settings.

The purpose of these controlled variants is to isolate the contribution of individual modules within the proposed framework. Each ablation removes exactly one component while keeping all remaining inference settings unchanged. This design allows us to attribute performance differences to specific modules rather than changes in decoding configuration or backbone models.

\begin{table*}[t]
\centering
\footnotesize
\begin{tabular}{p{3.2cm}p{2.0cm}p{9cm}}
\toprule
System & Type & Description \\
\midrule
MACAT & Ours & Multi-Agent with bridging, summary, knowledge injection, and boundary repair \\
w/o MC Bridge & Ablation & Removes classical-Chinese-to-modern-Chinese bridging \\
w/o Summary & Ablation & Removes local context summary \\
w/o RAG/Notes & Ablation & Removes rule-based knowledge and dynamic notes \\
w/o Replace Law & Ablation & Replaces the CLW knowledge base with out-of-domain legal knowledge \\
Google & Baseline & Commercial machine translation output \\
Model & Baseline & Direct translation from the base LLM \\
CoD & Strong baseline & External prompt-optimized system \\
Tower-plus & Strong baseline & Translation-specialized multilingual LLM baseline \\
\bottomrule
\end{tabular}
\caption{Systems and controlled variants. Each internal ablation changes one module while keeping the remaining inference settings fixed.}
\label{tab:systems}
\end{table*}
% \noindent\begin{minipage}{\columnwidth}
% \centering\footnotesize
% \resizebox{\columnwidth}{!}{%
% \begin{tabular}{p{2.3cm}p{1.3cm}p{4.3cm}}
% \toprule
% System & Type & Description \\
% \midrule
% MACAT & Ours & Multi-Agent with bridging, summary, knowledge injection, and boundary repair \\
% w/o MC Bridge & Ablation & Removes classical-Chinese-to-modern-Chinese bridging \\
% w/o Summary & Ablation & Removes local context summary \\
% w/o RAG/Notes & Ablation & Removes rule-based knowledge and dynamic notes \\
% w/o Replace Law & Ablation & Replaces the CLW knowledge base with out-of-domain legal knowledge \\
% Google & Baseline & Commercial machine translation output \\
% Model & Baseline & Direct translation from the base LLM \\
% CoD & Strong baseline & External prompt-optimized system \\
% Tower-plus & Strong baseline & Ranslation-specialized multilingual LLM baseline  \\
% \bottomrule
% \end{tabular}%
% }
% \captionof{table}{Systems and controlled variants. Each internal ablation changes one module while keeping the remaining inference settings fixed.}
% \label{tab:systems}
% \end{minipage}

\vspace{0.8em}
\section{Look up table for Chinese culture-loaded words}
\vspace{-0.3em}

\begin{figure*}[t]
\centering
\includegraphics[width=0.95\textwidth]{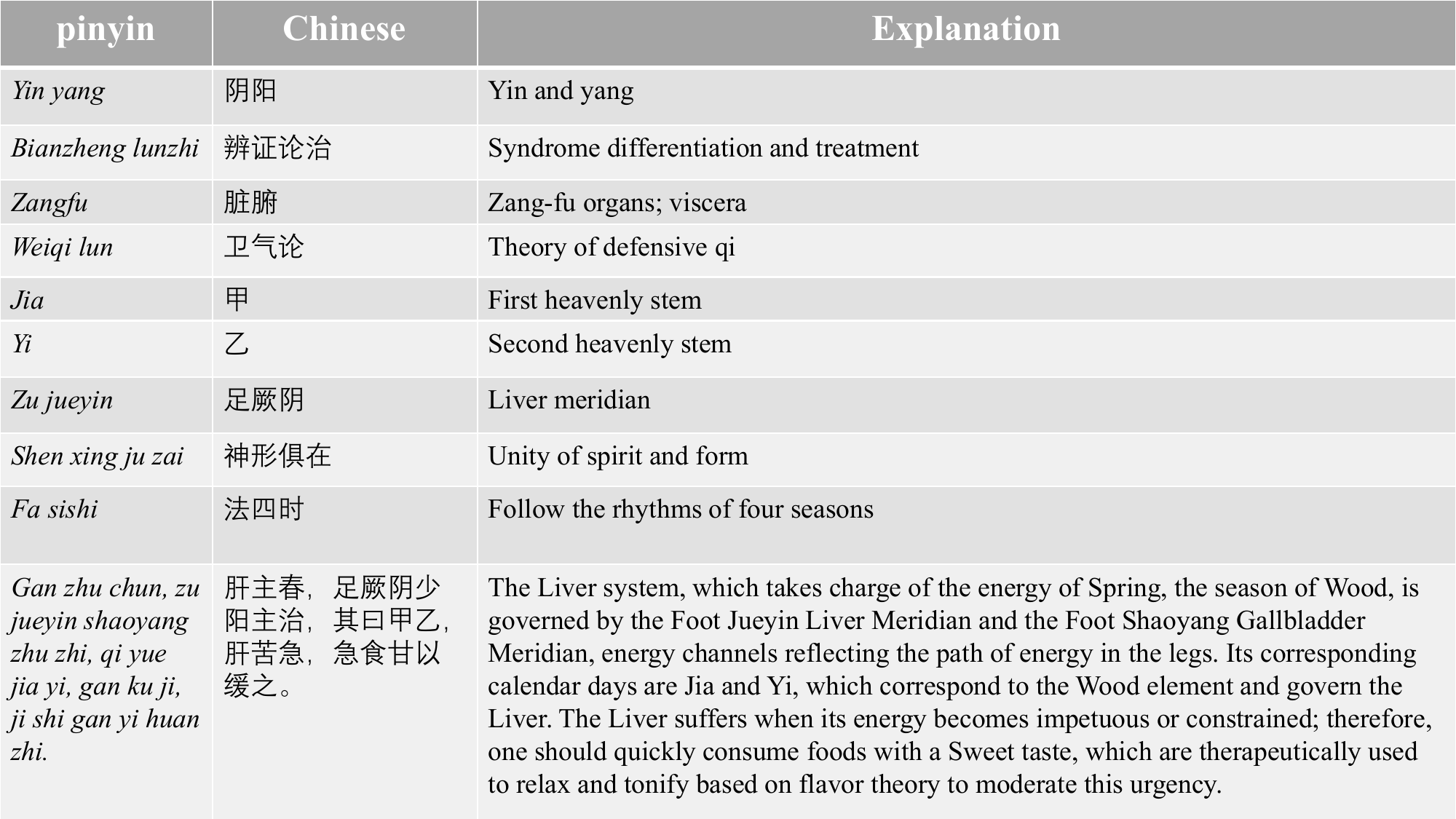}
\caption{Pinyin--Chinese--Definition lookup table for Chinese culture-loaded words appearing in this paper. Pinyin is shown without tone marks.}
\label{fig:pinyin-table}
\end{figure*}

Many culture-loaded words (CLWs) appearing in classical Chinese texts have no direct lexical equivalents in modern target languages. To improve readability and ensure terminological consistency, Figure~\ref{fig:pinyin-table} provides a lookup table containing representative CLWs discussed throughout this paper.

The table offers readers unfamiliar with Traditional Chinese Medicine (TCM) and classical Chinese culture a concise reference for understanding key concepts appearing in the examples and case studies. Besides, it illustrates the type of cultural knowledge that MACAT attempts to recover during translation, ranging from philosophical concepts and medical theories to culturally grounded entities such as Heavenly Stems and meridian systems.

%==============prompt==============
\vspace{0.8em}
\section{Prompt Templates} 
\vspace{-0.3em}
\label{sec:prompt-appendix}

This appendix provides the main prompt templates used by MACAT during inference. Since the proposed framework is entirely inference-time and does not require parameter fine-tuning, prompt design constitutes a central mechanism for coordinating the behavior of different agents.

The prompts are organized according to their functional roles in the pipeline, including knowledge base construction, dynamic culture-loaded word annotation, context processing, translation generation, and multi-dimensional evaluation.

All prompts are reported in their original form to facilitate reproducibility and provide a transparent description of the inference workflow.

\subsection{Knowledge Base Construction Prompts}
\label{sec:prompt-knowledge}

The prompts in this section are used to construct the rule-based cultural knowledge inventory employed by MACAT (See Table~\ref{tab:prompt-knowledge-extract}). Rather than extracting isolated facts, the objective is to identify reusable conceptual rules that capture recurring relationships among culture-loaded concepts.

The resulting rules are represented as structured knowledge cards containing trigger terms, conceptual mappings, application constraints, and generation templates. During translation, these rules are activated only when corresponding cultural triggers are detected in the source text, thereby supporting selective rather than unconditional knowledge injection.

\begin{table*}[t]
\centering
\footnotesize

\begin{tabular}{p{0.95\textwidth}}
\toprule
\textbf{prompt-knowledge-extract} \\
\midrule
\textbf{System:} ``You are an expert in Traditional Chinese Medicine (TCM) knowledge modeling.''

\textbf{User:} ``You will read an entire TCM classical document and extract ONLY high-value deep-logic knowledge rules. Before extracting rules, you MUST internally segment the document strictly by its chapters/sections. 

Selection constraints: Do NOT extract rules for every sentence. Only extract rules that generalize and will be reused. Prefer extracting rules around: Five Elements, Yin-Yang, Zang-Fu, channels/meridians, Qi/Blood/Essence/Spirit, seasonal correspondences, treatment principles, etiologies, pattern logic, organ-function relations. Each rule should have a small set of triggers (2--8). 

Output format: Output STRICT JSON: a single JSON array of objects. Each object MUST match: \{rule\_name, triggers, logic\_chain, application\_rule, output\_template\}. Placeholders like \{element\}, \{organ\} must be used. Do NOT use arrow symbols, parentheses, brackets, or asterisks in textual fields.'' \\
\bottomrule
\end{tabular}
\vspace{2mm}
\caption{Prompt for offline TCM knowledge rule extraction.}
\label{tab:prompt-knowledge-extract}
\end{table*}

\subsection{Dynamic CLW Annotation Prompts}
\label{sec:prompt-annotation}

While the rule-based inventory provides high-precision cultural knowledge, it cannot fully cover the diversity of CLWs encountered in historical texts. Therefore, MACAT additionally employs LLM-based dynamic annotation to identify culturally implicit concepts that are not explicitly represented in the predefined inventory.

The prompts in Tables~\ref{tab:prompt-annotation-auto} and \ref{tab:prompt-annotation-wordonly} support automatic CLW identification, explanatory note generation, keyword-guided annotation, and difficult-term discovery. These annotations are subsequently merged with rule-based knowledge and filtered through the knowledge budgeting mechanism.

\begin{table*}[t]
\centering
\footnotesize
\begin{tabular}{p{0.95\textwidth}}
\toprule
\textbf{prompt-annotation} \\
\midrule
\textbf{System:} ``You are a professional translator of classical Chinese texts.''

\textbf{User:} ``You are adding annotations to a Chinese classical text. Please annotate the culture-loaded Words in the following passage. Your annotations should help readers with no Chinese cultural background understand the text. The annotations should explain the traditional cultural concepts embedded in the classical text rather than merely paraphrasing the surface meaning. Keep the annotations concise. Output line by line. Only output the annotations and do not include any unrelated content. Do not fully repeat the original text. Follow the format shown in the example output.

Example Input: The Five Elements consist of metal, wood, water, fire, and earth...
Example Output:  
Five Elements: The five elemental categories and their generative and restrictive relationships...  
Metal, Wood, Water, Fire, Earth: Abstract functional attributes in Chinese cosmology...  
Five Zang Organs: The five functional organ systems including the heart, liver, spleen, lungs, and kidneys...
Input: \{para\}  
Output:'' \\
\midrule
\textbf{prompt-annotation-word-only} \\
\midrule
\textbf{System:} ``You are a professional Chinese-English translation expert.''

\textbf{User:} ``You are annotating a Chinese classical text. Identify the culture-loaded terms in the following passage. Output line by line. Only output the identified culture-loaded terms and do not include any unrelated content.

Example Input: The Five Elements consist of metal, wood, water, fire, and earth...

Example Output:  

Five Elements / Metal-Wood-Water-Fire-Earth / Five Zang Organs

Input: \{para\}  

Output:'' \\
\bottomrule
\end{tabular}
\vspace{2mm}
\caption{Prompts for LLM-based dynamic CLT annotation: auto-annotation and word-only extraction.}
\label{tab:prompt-annotation-auto}
\end{table*}

\begin{table*}[t]
\centering
\footnotesize
\begin{tabular}{p{0.95\textwidth}}
\toprule
\textbf{prompt-annotation-add-notes} \\
\midrule
\textbf{System:} ``You are an expert in classical Chinese translation and terminology standardization.''

\textbf{User:} ``You are adding annotations to a Chinese classical text. Based on the following passage, provide annotations for the given culture-loaded terms. 

Your annotations should help readers with no Chinese cultural background understand the text. The annotations should explain the traditional cultural concepts embedded in the classical text rather than merely paraphrasing the surface meaning. Keep the annotations concise. Output line by line. You may only annotate the given culture-loaded terms and must not arbitrarily add or remove terms. Only output the annotations and do not include any unrelated content.
Example Input: The Five Elements consist of metal, wood, water, fire, and earth...
Example Output:  
Five Elements: The five elemental categories and their generative and restrictive relationships...
Input:  
Passage: \{para\}  
Culture-Loaded Terms: \{keywords\}  
Output:'' \\
\midrule
\textbf{prompt-find-difficult-words} \\
\midrule
\textbf{System:} ``You are a professional Chinese-English translation expert.''

\textbf{User:} ``You are preparing annotations for a Chinese classical text passage. Identify the proper nouns and historical or cultural allusions that are unique to ancient Chinese culture. The identified terms should not be words that are still commonly used in modern Chinese. The identified terms must be proper nouns. Output only the identified terms, one per line, without any additional explanation.
Input: \{input\}  
Output:'' \\
\bottomrule
\end{tabular}
\vspace{2mm}
\caption{Additional annotation prompts: keyword-guided annotation to add notes and difficult-term identification.}
\label{tab:prompt-annotation-wordonly}
\end{table*}

\subsection{Context Processing Prompts}
\label{sec:prompt-context}

Context processing aims to improve translation consistency without introducing additional cultural knowledge. MACAT incorporates two auxiliary context-conditioning mechanisms: local summarization and Classical Chinese bridging.

Local summarization compresses neighboring segments into a concise contextual representation (See Table~\ref{tab:prompt-summary}), helping preserve topical continuity and previously introduced explanations. Classical Chinese bridging rewrites Classical Chinese into modern Chinese while retaining culture-loaded expressions, thereby reducing linguistic ambiguity before translation.

In addition, Table~\ref{tab:prompt-margin} presents the boundary-aware repair prompt. This component operates after candidate selection and focuses on correcting inconsistencies across adjacent segments while preserving previously generated content.

\begin{table*}[t]
\centering
\footnotesize
\begin{tabular}{p{0.95\textwidth}}
\toprule
\textbf{prompt-summary} \\
\midrule
\textbf{System:} ``You are a professional Chinese-English translation expert.''

\textbf{User:} ``This is a passage summarization task. Please summarize the following excerpt from a Chinese classical text. In your summary, briefly describe the main content of the passage. Only output the summary itself and do not include any additional content.
\{corpus\}
Summarize the corpus.'' \\
\midrule
\textbf{prompt-translation-mc} \\
\midrule
\textbf{System:} ``You are an expert in translating Classical Chinese into Modern Chinese.''

\textbf{User:} ``This is a Classical Chinese to Modern Chinese translation task. Please translate the following passage into Modern Chinese. During translation, you must use modern Chinese vocabulary, grammar, and usage conventions. Preserve all $<$n$>$$\dots$$<$/n$>$ tags exactly as they appear, and do not merge or split lines. Output only the translated text without any explanation.
Source Text: \{source\}'' \\
\bottomrule
\end{tabular}
\vspace{2mm}
\caption{Context processing prompts: summarization and classical Chinese bridging.}
\label{tab:prompt-summary}
\end{table*}

\begin{table*}[t]
\centering
\footnotesize
\begin{tabular}{p{0.95\textwidth}}
\toprule
\textbf{prompt-margin} \\
\midrule
\textbf{System:} ``You are a professional Chinese-English translation expert.''

\textbf{User:} ``This is a translation task. Please translate the following passage into English. The given passage contains partially translated English sentences and partially untranslated Chinese sentences. Under the premise of maintaining semantic coherence, translate the entire passage into English.

During translation, you must follow standard English vocabulary, grammar, and usage conventions. Only output the translation result without any additional explanation. The output should contain the complete translated passage, including both the originally existing English content and the newly translated Chinese content.
When generating the output, preserve the line numbering format from the input (e.g., $<$0$>$text$<$/0$>$ indicates the first line), and ensure that the number of output lines exactly matches the number of input lines.
The original text is as follows:
\{upper\}
\{middle\}
\{lower\}
English Translation:'' \\
\bottomrule
\end{tabular}
\vspace{2mm}
\caption{Boundary-aware merge and repair prompt.}
\label{tab:prompt-margin}
\end{table*}

\subsection{Translation Generation Prompts}
\label{sec:prompt-translation}

The prompt in this section is responsible for generating translation candidates under different knowledge-conditioning settings. It serves as the primary translation interface of MACAT. See Table~\ref{tab:prompt-trans-en}.

The design follows the principle of controlled explicitation. Background knowledge is provided only when necessary for understanding culturally opaque concepts, while excessive explanatory expansion is explicitly discouraged. By keeping the instruction template fixed and varying only the injected knowledge and contextual signals, MACAT ensures that performance differences arise from information availability rather than prompt engineering artifacts.

\begin{table*}[t]
\centering
\footnotesize
\begin{tabular}{p{0.95\textwidth}}
\toprule
\textbf{prompt-translation-en} \\
\midrule
\textbf{System:} ``You are a professional translator of Huangdi Neijing (Traditional Chinese classics).''

\textbf{User:} ``This is a Chinese to English translation task. Your goal is `Explicit Translation', making the TCM terminology and logic understandable for readers without a Chinese background. 

Primary goal: communicative and culturally faithful translation that explains implicit logic. Constraints (MUST follow): 

(1) The input lines may include tags like $<$0$>$$\dots$$<$/0$>$. Preserve ALL such tags exactly. (2) Keep the SAME number of lines as the input. Do NOT merge, split, reorder, or drop lines. (3) Use the provided background knowledge (footnote) only when it is necessary for reader comprehension of opaque TCM terms or cultural logic. (4) Do NOT just provide a literal translation; ensure the reader understands WHY terms like `Jia' or `Yi' are used (e.g., their connection to the Liver and Wood element). (5) Use consistent, standard renderings for key TCM terms (Qi, Yin-Yang, Zang-Fu, channels/meridians, syndrome patterns, formula names). (6) If the provided footnotes contain guidance for a term, prefer and follow them. (7) Do NOT paraphrase all notes into the translation. (8) Prefer brief apposition or a short parenthetical clarification only for the 1--2 truly opaque terms. (9) If no clarification is necessary, keep the translation concise. Your output should have the same format as the input.
Source: \{source\} Background Knowledge and Logic Chain: \{footnote\}'' \\
\bottomrule
\end{tabular}
\vspace{2mm}
\caption{Primary translation generation prompt.}
\label{tab:prompt-trans-en}
\end{table*}

\subsection{Evaluation Prompts}
\label{sec:prompt-evaluation}

This section presents the prompts used by the multi-dimensional evaluation agent. The evaluation framework follows the three-round protocol introduced in Table~\ref{tab:prompt-eval-rounds}, consisting of initial assessment, reflective reassessment, and reference-assisted factual verification.

Each round evaluates translations along five dimensions: Terminological Precision, Readability, Fidelity, Cultural Preservation, and Cultural Explicitation. The prompts are designed to encourage explicit reasoning about culturally grounded translation quality and to reduce common evaluation biases such as halo effects, scale inconsistency, and length preference.

Reporting the complete evaluation prompts improves transparency and facilitates future comparison with alternative LLM-based evaluation protocols.

\begin{table*}[t]
\centering
\footnotesize
\setlength{\tabcolsep}{6pt}
\renewcommand{\arraystretch}{1.2}
\begin{tabular}{p{0.22\textwidth}p{0.72\textwidth}}
\toprule
\textbf{Prompt-Evaluation} & \textbf{Detailed Criteria} \\
\midrule

Prompt-Round1 \newline
&
\textbf{System:} ``You are a senior evaluator for English translations of Traditional Chinese Medicine classics.''

\vspace{0.3em}

\textbf{User:} ``You are an expert evaluator of translations for Chinese cultural and classical texts. Please evaluate the translation across five dimensions on a 1--10 integer scale, and provide one concise justification for each dimension.

Evaluation dimensions:
(1) Terminological Precision,
(2) Cultural Fidelity,
(3) Syntactic Fluency,
(4) Literal Information Completeness,
(5) Implicit Knowledge Explicitation.

Scoring criteria:
1 = core terminology is entirely mistranslated or omitted;
5 = generally correct but inconsistent or non-standard;
10 = all terminology adopts academically recognized English renderings and accurately preserves semantic relations.

Output requirements:
Only output a single JSON object using the following keys:
terminology\_accuracy,
grammar\_readability,
cultural\_imagery,
literal\_information\_completeness,
implicit\_knowledge\_explicitation.
Source text: \{source\}
Translation: \{translation\}''
\\
\midrule
Prompt-Round2 \newline
&
\textbf{System:} ``You are a senior evaluator for English translations of Traditional Chinese Medicine classics.''

\vspace{0.3em}

\textbf{User:} ``Below is your evaluation result from the previous round for the same source text and translation (JSON): \{prev\_json\}

Please carefully review your previous assessment.

Determine whether any of the following issues are present:

- Inconsistency across evaluation dimensions (e.g., poor readability but unusually high style-related scores);
- Judgments on terminology or cultural interpretation that deviate from academically accepted standards;
- Overly lenient or overly harsh scoring;
- Halo Effect: assigning overly positive scores because the translation appears generally fluent while overlooking critical terminology errors;
- Length Bias: unfairly penalizing a translation for being longer, or rewarding it merely for being shorter;
- Scale Mismatch: inconsistent scoring standards across the four evaluated versions.

Please re-examine both the source text and the translation, then output the revised scores together with revision rationales.

If no revision is necessary, explicitly confirm and output the same scores.

Output format requirements (must be strictly followed):

- Only output a JSON object;
- Do not output any explanatory text;
- The JSON structure must remain exactly the same as in the previous round;'' 
Source text: \{source\}
Translation: \{translation\}''
\\
\midrule
Prompt-Round3 \newline
&
\textbf{System:} ``You are a senior evaluator for English translations of Traditional Chinese Medicine classics.''

\vspace{0.3em}

\textbf{User:} ``Below is an expert reference translation for the same source passage (not a gold-standard answer, but provided only for factual verification):
\{reference\}

Do not treat the reference translation as the upper bound of expression quality.
Also, do not penalize the system translation simply because it is more explicit or explanatory than the reference translation.

You should use the reference translation only for the following two verification purposes:

(1) Semantic Alignment:
Does the system translation convey the same core cultural facts as the reference translation?

(2) Gain Analysis:
Are the additional explanatory contents in the system translation instances of
Value-added Elaboration
(necessary and accurate explicitation),
or
Redundant Paraphrase
(unnecessary rewriting or over-elaboration)?

Scoring principles:

If the system translation is more explicit than the reference translation, but the added explanations are culturally accurate, necessary, and concise, it should not be penalized.

If the system translation is merely shorter or smoother, but makes the text harder for readers without a TCM background to understand, it should not receive additional credit.

Output requirements:
Only output a JSON object.
The JSON structure must remain exactly the same as in the previous round.

Source text: \{source\}

Translation: \{translation\}''
\\

\bottomrule
\end{tabular}
\caption{Evaluation prompts for Round1 to Round3.}
\label{tab:prompt-eval-rounds}
\end{table*}

\end{document}